\documentclass[conference, onecolumn]{IEEEtran}
\IEEEoverridecommandlockouts
\usepackage{cite}
\usepackage{amsmath,amssymb,amsfonts}
\usepackage{multirow}
\usepackage{subfigure}
\usepackage{enumerate}
\usepackage{algorithmic}
\usepackage{graphicx}
\usepackage{textcomp}
\usepackage{xcolor}
\def\BibTeX{{\rm B\kern-.05em{\sc i\kern-.025em b}\kern-.08em
    T\kern-.1667em\lower.7ex\hbox{E}\kern-.125emX}}
\begin{document}

\title{Lung CT Imaging Sign Classification through Deep Learning on Small Data
}

\author{\IEEEauthorblockN{Guocai He}
}

\maketitle

\begin{abstract}
The annotated medical images are usually expensive to be collected. This paper proposes a deep learning method on small data to classify Common Imaging Signs of Lung diseases (CISL) in computed tomography (CT) images. We explore both the real data and the data generated by Generative Adversarial Network (GAN) to improve the reliability and the generalization of learning. First, we use GAN to generate a large number of CISLs from small annotated data, which are difficult to be distinguished from real counterparts. These generated samples are used to pre-train a Convolutional Neural Network (CNN) for classifying CISLs. Second, we fine-tune the CNN classification model with real data. Experiments were conducted on the LISS database of CISLs. We successfully convinced radiologists that our generated CISLs samples were real for 56.7\% of our experiments. The pre-trained CNN model achieves 88.4\% of mean accuracy of binary classification, and after fine-tuning, the mean accuracy is significantly increased to 95.0\%. For multi-classification of all types of CISLs and normal tissues, through the two stages of training, the mean accuracy, sensitivity and specificity are up to about 91.83\%, 92.73\% and 99.0\%, respectively. To our knowledge, this is the best result achieved on the LISS database, which demonstrates that the proposed method is effective and promising for fulfilling deep learning on small data.
\end{abstract}
\begin{IEEEkeywords}
lung CT imaging signs classification, deep learning, small data, Generative Adversarial Network (GAN), Convolutional Neural Networks (CNNs)
\end{IEEEkeywords}

\section{Introduction}
Lung cancer is a leading type of human cancers with a high mortality. Detection of lung diseases is an important task of cancer prevention and treatment and an essential problem in medical diagnosis. Computed tomography (CT) image is a crucial factor for medical diagnosis of lung diseases, and the automatic classification of CT images is an important application of the computer aided diagnosis (CAD) system. Pulmonary nodules are spots on the lung that usually reflect lung illnesses, and numerous efforts \cite{ye2009shape,li2007recent} have been made on lung CT imaging analysis. These contributions are mainly concerned with lung nodules. However, not all lung diseases are presented as pulmonary nodules, the lung CT imaging signs are more crucial information in the diagnosis of lung diseases \cite{han2015liss}. The radiologists make decisions based on the correlation between CT imaging signs and lung diseases. Therefore, in this paper, we contribute to classify Common Imaging Signs of Lung diseases (CISLs).

As we know, deep learning method is one of the most effective way to solve the task of image classification currently. However, in the field of medical imaging, one of the major challenges at present is the lack of large-scale well-annotated datasets of medical images. Training deep classification models requires a large amount of annotated data. Unlike natural images, high-quality annotation by radiology expert is often costly. To deal with such problem, the main two techniques are data augmentation and transfer learning. In traditional data augmentation method \cite{krizhevsky2012imagenet}, a variety of visual transformations of an image are used to yield images, such as rotation, noise, shift, distort, zoom in/out, flip, shear, and so on. Such strategy can introduce the diversity into the training data to some extent and reduce the risks of overfitting of the models. However, the diversity brought by visual transformation is still not enough, besides, these transformations affect the distribution of the original dataset and introduce unnecessary noises and offset. Furthermore, when dealing with different datasets, we need to get familiar with the target dataset, and decide which image transformations to be introduced. This process always requires human involvements and thus the traditional data augmentation is low scalable. The transfer learning is to performing pre-training on large-scale natural images dataset (i.e. ImageNet \cite{russakovsky2015imagenet}) and then fine-tuning on target dataset. These models pre-trained on ImageNet dataset can provide rich hierarchical image features. Recently, CNNs pre-trained on ImageNet have been used for thoraco-abdominal lymph node (LN) detection and interstitial lung disease (ILD) classification \cite{shin2016deep}. However, the limited target dataset is still a bottleneck to performance of transfer learning, because fine-tuning CNN models on limited data may easily lead to overfitting.

With the introduction of Generative Adversarial Networks (GANs) \cite{goodfellow2014generative}, the generative learning has gained more attention in the fields of computer vision and natural language processing. For medical imaging, there are also some applications of GAN, such as low dose CT denoising \cite{wolterink2017generative}, segmentation \cite{zhang2017deep}, detection \cite{schlegl2017unsupervised}, reconstruction \cite{shitrit2017accelerated} and image synthesis \cite{bayramoglu2017towards}. GAN can learn to generate high-quality images, achieve cross-modal data generation and image-to-image transformation. Maria et al. \cite{chuquicusma2017fool} contributed to generate lung nodules using GAN, but the generated nodules they showed were derived from a few patterns, i.e., the model fell into the dilemma of overfitting.

In this paper, we propose a GAN-based method to learn the classifier of Common Imaging Signs of Lung diseases over small data. The core idea is that we generate CISLs based on GAN and combine these generated samples and real data to train the CNN classifier. Unlike the method in \cite{frid2018synthetic} that uses the traditional data augmentation to extend the training data of the GAN model and trains the liver lesion classifier with the mixture of the generated data and real data, we train the GAN with only the real data and adopt the two-stages training scheme (i.e. pre-training with generated data and fine-tuning with real data). First, we use CISLs samples to train CVAE-GAN \cite{bao2017cvae}. Second, we use the well-trained model of GAN to generate a large number of samples and feed these samples into the CNN classifier for pre-training. Finally, we use the real CISLs samples to fine-tune the pre-trained CNN classifier. Although the two-stages training procedure looks like the strategy of transfer learning, but our method is different in that it does not exploit additional knowledge from other datasets, which is different from that of transfer learning. The proposed method is evaluated on the LISS \cite{han2015liss} database of CISLs.

We make the following three contributions:
\begin{enumerate}[\hspace{0.5cm}1)]
\item We generate high-quality CISLs samples based on CVAE-GAN.
\item We pre-train CNN classifier with the generated imaging signs and fine-tune CNN with real CISLs, and we improve the performance of CNN classifier of CISLs, particularly when the number of available training samples is limited.
\item We show that the proposed approach outperforms the methods of traditional data augmentation and transfer learning from pre-trained ImageNet models on the limited medical images database.
\end{enumerate}
\section{The Proposed Method}
\begin{figure}
\centering
\includegraphics[width=0.5\paperwidth]{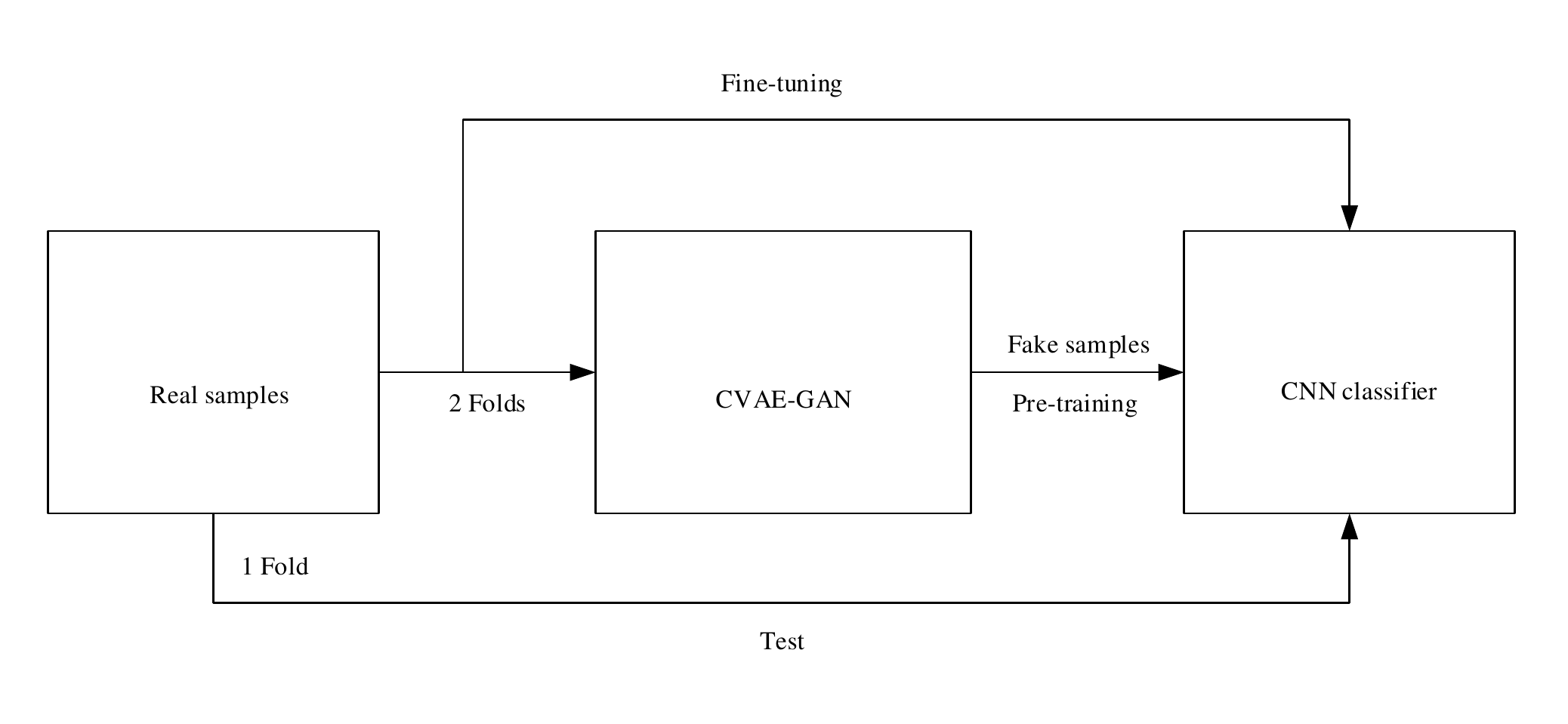}
\caption{Overall framework of the proposed method: using 2 folds of real samples to train CVAE-GAN, and generating CISLs to pre-train CNN classifier, then fine-tuning CNN model with these 2 folds of samples.}
\label{fig:framework}
\end{figure}

The training of convolutional neural network usually requires a large number of semantic labeled data. It is always difficult to train a convolutional neural network classifier on a limited set of samples. Even for a simple convolutional neural network, the number of the parameters still exceeds million. Training on small data sets is easy to lead to overfitting and poor generalization performance. In order to make better use of the spontaneous and efficient capability for feature extraction of convolutional neural networks, we train GAN model to generate CISL samples which follow the probability distribution of the original dataset, then we extend the training data of classification model and use two-stages scheme to improve the adaptability of CNN classifier on small datasets. The overall framework of our method is illustrated in Fig.~\ref{fig:framework}, and the details of it are described in the following.

\subsection{GANs and Data Generation}
The basic GAN model consists of a generator network G and a discriminator network D. During training, the network G is responsible for generating samples which are confused with real samples so that the network D is not able to easily distinguish fake samples from real samples. At the same time, the network D learn to distinguish real samples from fake ones as much as possible. The optimization objective can be simply defined as:
\begin{align}
\min\limits_{G}\max\limits_{D}{V(G,D)}=\mathbb{E}_{x_{\sim}p_{data}}[log(D(x))]+
\mathbb{E}_{z_{\sim}p_{z}}[log(1-D(G(z)))],
\end{align}
where $z$ is the prior input to the G network and is usually sampled from gaussian distribution or uniform distribution, $x$ denotes the real data which is sampled from the original dataset.

The basic GAN has strong randomness and is difficult to be trained. The two main problems in GAN training are mode collapse and instability of learning. In order to reduce the search space, conditional GAN (CGAN) \cite{mirza2014conditional} introduces conditional information to GAN, which can be the semantic label of the sample, the image or other modal data, but the problems have not been solved well yet. CVAE-GAN has been validated to be a good solution to the two problems. This model combines two generative models CVAE \cite{kingma2014semi} and CGAN, and each of them has different advantages. CVAE has the advantage of reducing the risks of mode collapse, and GAN is beneficial to generate higher quality images.

\begin{figure}
\centering
\includegraphics[width=0.5\paperwidth]{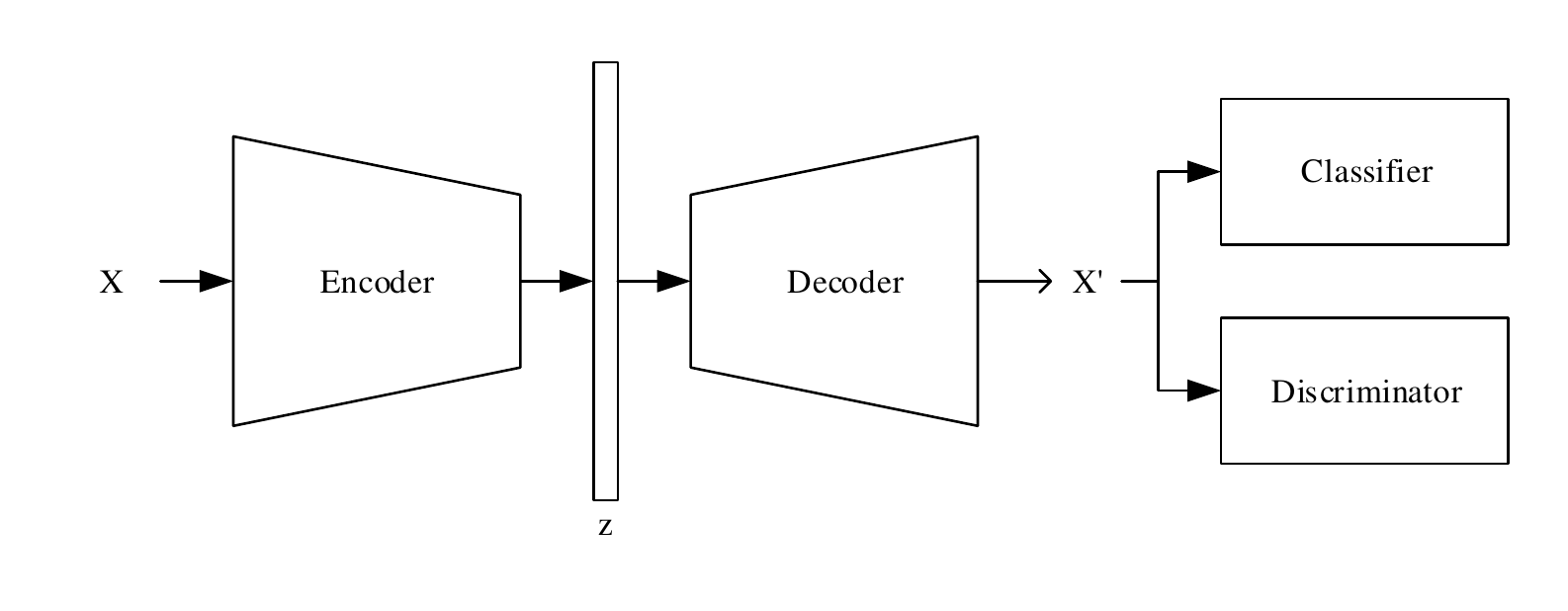}
\caption{The CVAE-GAN architecture used for generating samples of CISLs.}
\label{fig:GAN}
\end{figure}

We use CVAE-GAN to generate all kinds of CISLs. Our proposed model is shown in Fig.~\ref{fig:GAN}. The encoder module encodes the samples $X$ from original dataset, and outputs combination of the mean $\mu$ and the covariance $\epsilon$ of the latent variables. Then we sample the latent variables $z=noise\times{log(\epsilon)}+\mu$, which is the input to decoder module, where noise$\sim$Gaussian (0, 1). The decoder module thus learns to generate realistic samples $X'$, and the discriminator should distinguish between $X$ and $X'$ as much as possible. The classifier module is trained using the real samples and the corresponding labels. When the synthetic samples are taken as inputs of classifier and discriminator, we use mean feature matching, pair-wise feature matching and the $L_2$ reconstruction loss to provide extra gradients to the CVAE to help generate more plausible CISLs. In this paper, we simply choose the input of the last fully connected layer of the classifier module and discriminator module as the data for feature matching.
\subsection{CISLs Classification}
The LISS \cite{han2015liss} database explicitly contains 9 types of CISLs, and the details of it are described in Section \ref{dataset}. We categorized the classifications of CISLs into binary classification and multi-classification. For the binary classification, we classified sign regions and non-sign regions separately for the nine CISLs (e.g. GGO and non-GGO). For the multi-classification, we integrated the nine CISLs and normal tissue for the classification of 10 categories. We conducted the classification experiments from the two aspects, respectively.

For all experiments of classification, the CNN classifiers consist of 4 convolutional layers and 2 fully connected layers. And for the binary classification and multi-classification, the classifier has just 2 and 10 neurons in the last fully connected layer, respectively. The layers of the classification neural network are listed below. Obviously, the architecture of the network we used is not very essential for the strength of classification.
\begin{enumerate}[\hspace{0.5cm}(1)]
\item Convolution layer with 5$\times$5 filters, stride 1, padding SAME, 32 output channels. ReLU activation function.
\item Convolution layer with 5$\times$5 filters, stride 2, padding SAME, 64 output channels. Batch normalization, ReLU activation function.
\item Convolution layer with 5$\times$5 filters, stride 2, padding SAME, 128 output channels. Batch normalization, ReLU activation function.
\item Convolution layer with 5$\times$5 filters, stride 2, padding SAME, 256 output channels. Batch normalization, ReLU activation function.
\item Fully Connected layer with 1024 output neurons. Batch normalization, ReLU activation function, dropout 0.5.
\item Fully Connected layer with 2/10 (2 for binary classification, 10 for multi-classification) output neurons. Softmax activation function.
\end{enumerate}

\section{Experiments}
\subsection{Dataset\label{dataset}}
In this paper, we used the publicly available CISLs dataset, the LISS, to perform our experiments. The dataset contains 9 types of CISLs, including Ground Glass Opacity (GGO), lobulation, calcification, Cavity \& Vacuolous (CV), spiculation, Pleural Indentation (PI), Bronchial Mucus Plugs (BMP), Air Bronchogram (AB), and Obstructive Pneumonia (OP). The dataset contains about 25800 lung CT images, and the size of image is 512 $\times$ 512. Radiologist adopted rectangular bounding box to mark the region of interest (ROI). Because the areas of CISLs in the dataset are not the same size, we unified the size of CISLs to 64$\times$64 for simplicity. For CISL whose region size is less than 64, we included the background around the lesion region into the sample, placing the rectangle box of the CISL to the center of the sample. For CISL whose region size is larger than 64, we used sliding window and image scaling by bilinear interpolation algorithm to yield samples. For CISLs whose regions cover nearly half of the CT images, considering these samples are not well-annotated, we excluded them directly. We totally collected eligible 696 CISLs, including 205 GGOs, 41 lobulations, 47 calcifications, 167 CVs, 29 spiculation, 80 PIs, 86 BMPs, 23 ABs and 18 OPs.

\subsection{Implementation Details}
All CISLs data is normalized to -1$\sim$1 by Min-Max normalization technique. We use CVAE-GAN to learn to generate samples of CISLs, and the architecture of each module of CVAE-GAN is the following.

The architecture of Encoder in CVAE-GAN:
\begin{enumerate}[\hspace{0.5cm}(1)]
\item Convolution layer with 5$\times$5 filters, stride 2, padding SAME, 32 output channels. LReLU activation function (leak=0.2).
\item Convolution layer with 5$\times$5 filters, stride 2, padding SAME, 64 output channels. Batch normalization, LReLU activation function (leak=0.2).
\item Convolution layer with 5$\times$5 filters, stride 2, padding SAME, 128 output channels. Batch normalization, LReLU activation function (leak=0.2).
\item Convolution layer with 5$\times$5 filters, stride 2, padding SAME, 256 output channels. Batch normalization, LReLU activation function (leak=0.2).
\item Fully Connected layer with 1024 output neurons. Batch normalization, LReLU activation function (leak=0.2).
\item Fully Connected layer with 128 output neurons. Batch normalization, LReLU activation function (leak=0.2).
\end{enumerate}

The architecture of Decoder in CVAE-GAN:
\begin{enumerate}[\hspace{0.5cm}(1)]
\item Fully Connected layer with 1024 output neurons. Batch normalization, LReLU activation function (leak=0.2).
\item Fully Connected layer with 4096 output neurons. Batch normalization, LReLU activation function (leak=0.2).
\item Transposed Convolution layer with 5$\times$5 filters, stride 2, padding SAME, 128 output channels. Batch normalization, LReLU activation function (leak=0.2).
\item Transposed Convolution layer with 5$\times$5 filters, stride 2, padding SAME, 64 output channels. Batch normalization, LReLU activation function (leak=0.2).
\item Transposed Convolution layer with 5$\times$5 filters, stride 2, padding SAME, 32 output channels. Batch normalization, LReLU activation function (leak=0.2).
\item Transposed Convolution layer with 5$\times$5 filters, stride 2, padding SAME, 1 output channels. Batch normalization, tanh activation function.
\end{enumerate}

The architecture of Discriminator in CVAE-GAN:
\begin{enumerate}[\hspace{0.5cm}(1)]
\item Convolution layer with 5$\times$5 filters, stride 1, padding SAME, 32 output channels. Batch normalization, ReLU activation function.
\item Convolution layer with 5$\times$5 filters, stride 2, padding SAME, 64 output channels. Batch normalization, ReLU activation function.
\item Convolution layer with 5$\times$5 filters, stride 2, padding SAME, 128 output channels. Batch normalization, ReLU activation function.
\item Fully Connected layer with 512 output neurons. Batch normalization, ReLU activation function.
\item Fully Connected layer with 1 output neuron. Sigmoid activation function.
\end{enumerate}

The architecture of Classifier in CVAE-GAN:
\begin{enumerate}[\hspace{0.5cm}(1)]
\item Convolution layer with 5$\times$5 filters, stride 1, padding SAME, 32 output channels. Batch normalization, ReLU activation function.
\item Convolution layer with 5$\times$5 filters, stride 2, padding SAME, 64 output channels. Batch normalization, ReLU activation function.
\item Convolution layer with 5$\times$5 filters, stride 2, padding SAME, 128 output channels. Batch normalization, ReLU activation function.
\item Fully Connected layer with 512 output neurons. Batch normalization, ReLU activation function.
\item Fully Connected layer with 9 output neurons. Softmax activation function.
\end{enumerate}

In all experiments, we used 3-fold cross-validation. The CVAE-GAN and all CNN classifiers are implemented with TensorFlow framework, and these deep networks are trained using an NVIDIA GeForce GTX 1080 GPU.

\subsection{Results}
The data generated by the GAN is de-normalized to yield the final CISL samples. Then, we will present both the quality of the generated samples and the performance of classification in following sub-sections.

\subsubsection{Quality of the Generated CISLs}
\begin{figure}[t]
\centering
\includegraphics[width=0.5\paperwidth]{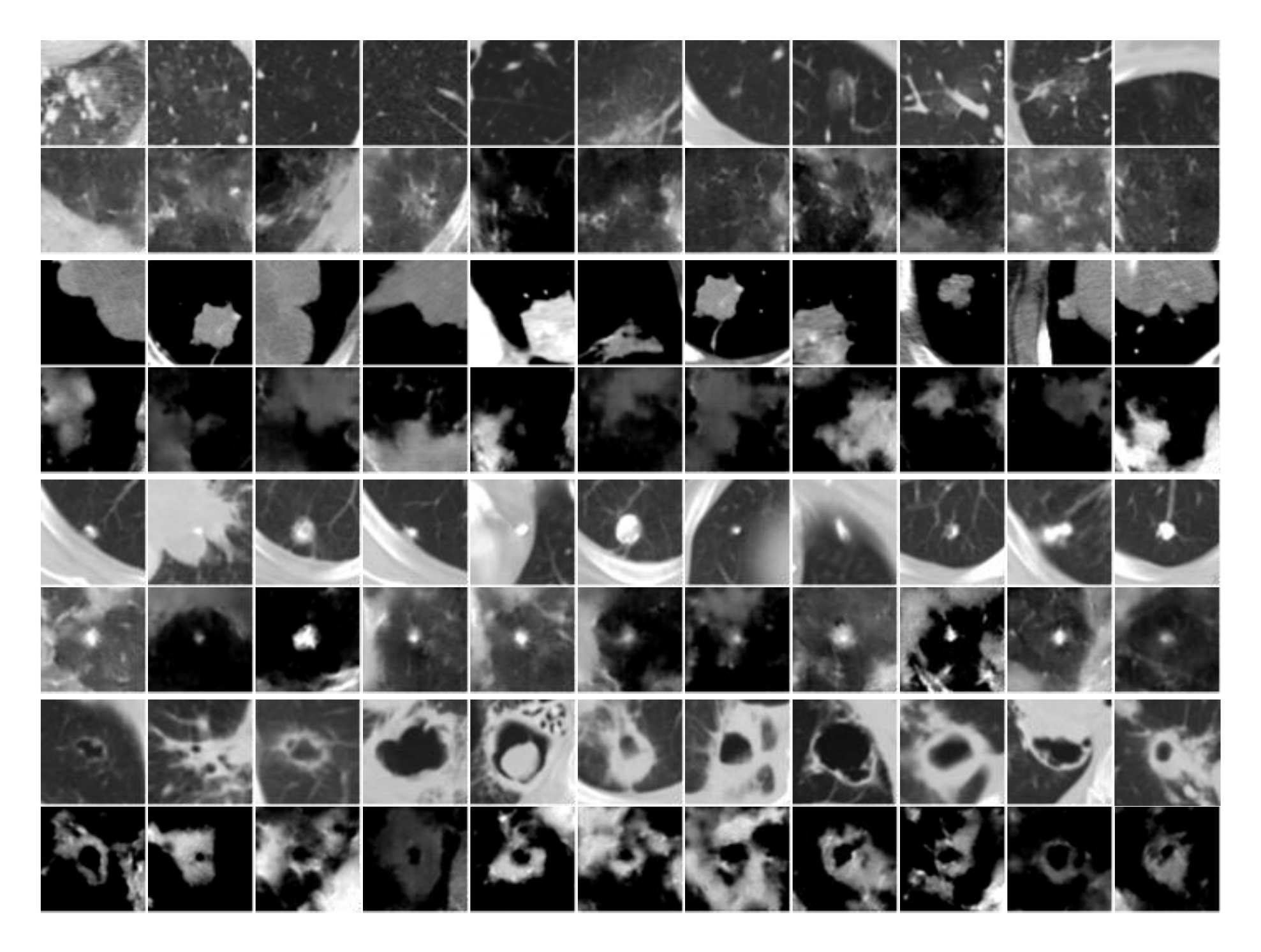}
\caption{Visual comparison of real and generated CISLs. Every two rows correspond to different types of CISLs from top to bottom (GGO, lobulation, calcification and CV) and in each two rows, samples of the first row are real and samples of the second row are generated by GAN.}
\label{fig:generated samples}
\end{figure}

The generated CISLs are presented in Fig.~\ref{fig:generated samples}. The synthetic images are similar to the real samples in appearance. And almost all of these generated samples are different from each other. The generated samples belonging to same category have quite different characteristics. It can be seen that the proposed algorithm can effectively avoid mode collapse and generate realistic samples of CISLs.

We invited two experts with 17 and 14 years of experience of reading lung CT images in radiology department to participate in the visual Turing Test of the quality of the generated CISLs. For each type of CISL, we randomly selected 10 real and 10 fake samples, respectively. Then we showed these 180 samples of CISLs to the two radiologists and informed them that these images may be three possible combinations (all real, all fake or mixture of the real and the fake). The two radiologists needed to independently judge whether each sample is real or fake, then we collected the experiment results. The radiologist with 17 years of professional experience wrongly took 46.7\% of the generated samples as real and 26.7\% of the real samples as fake. The another radiologist with 14 years of professional experience misidentified 66.7\% of the fake samples as real and 46.7\% of the real samples as fake. These experiment results indicate that these generated CISLs are high-quality and can be used for improving the performance of CISLs classification.
\subsubsection{Binary Classification}
We use the CNN as a classifier and pre-train it with the generated samples. For each type of CISLs, we randomly sample 18000 from the pool of generated data. For non-sign samples, accordingly, we randomly cut the same number of 64$\times$64 samples from the CT scans which do not contain the corresponding kind of lung signs. After the pre-training procedure is completed, we test the pre-trained model using the real samples of CISLs. And in test dataset, we still balance the numbers of the signs (1 fold) and non-signs. We compare proposed method with the traditional method of data augmentation. For the traditional data augmentation, we adopted 5 kinds of visual transformations for all types of CISLs:
\begin{enumerate}[\hspace{0.5cm}(a)] 
\item Rotation. Degree range for random rotations is -$30\,^{\circ}$$\sim$ $30\,^{\circ}$.
\item Randomly shift horizontally/vertically. Range is -0.1 $\sim$ 0.1.
\item Shear. The shear angle is about -$10\,^{\circ}$$\sim$ $10\,^{\circ}$.
\item Zoom in/out. Range for random zoom is 0.9 $\sim$ 1.1.
\item Randomly flip samples horizontally/vertically.
\end{enumerate}
Similarly, we used the data generated by above visual transformations to train CNN classifier and tested using real samples. To be fair, the number of samples yielded by the traditional method of data augmentation is same as the proposed method. In order to evaluate the performance of the trained models, we used three indicators as the metrics of the classifications, i.e., accuracy (AC), sensitivity (SE) and specificity (SP). The experiment results are shown in Table~\ref{table:table1}.

\setlength{\tabcolsep}{5pt}
\begin{table}
\begin{center}
\caption{Model’s performance of binary classification for each type of CISLs.}
\label{table:table1}
\begin{tabular}{cccccccccc}
\hline\noalign{\smallskip}
\multirow{2}{*}{CISLs} & \multicolumn{3}{c}{Pre-training} & \multicolumn{3}{c}{Fine-tuning} & \multicolumn{3}{c}{Traditional method}\\
\cline{2-10}\noalign{\smallskip}
  & AC & SE & SP & AC & SE & SP & AC & SE & SP\\
\noalign{\smallskip}
\hline
\noalign{\smallskip}
GGO  & 0.904 & 0.736 & 0.988 & 0.926 & 0.919 & 0.932 & 0.901 & 0.951 & 0.851\\
lobulation & 0.966 & 0.931 & 1.000 & 0.983 & 1.000 & 0.967 & 0.859 & 0.785 & 0.933\\
calcification & 0.913 & 0.826 & 1.000 & 0.946 & 0.927 & 0.967 & 0.868 & 0.781 & 0.956\\
CV & 0.866 & 0.732 & 1.000 & 0.949 & 0.940 & 0.955 & 0.960 & 0.942 & 0.978\\
spiculation & 0.860 & 0.760 & 0.960 & 0.940 & 0.919 & 0.964 & 0.836 & 0.671 & 1.000\\
PI & 0.882 & 0.764 & 1.000 & 0.965 & 0.944 & 0.986 & 0.972 & 0.944 & 1.000\\
BMP & 0.826 & 0.681 & 0.971 & 0.920 & 0.912 & 0.927 & 0.920 & 0.927 & 0.913\\
AB & 0.886 & 0.909 & 0.864 & 0.955 & 0.962 & 0.961 & 0.836 & 0.720 & 0.952\\
OP & 0.857 & 0.714 & 1.000 & 0.964 & 0.944 & 1.000 & 0.800 & 0.600 & 1.000\\
mean & 0.884 & 0.784 & 0.976 & 0.950 & 0.941 & 0.962 & 0.883 & 0.813 & 0.954\\
\hline
\end{tabular}
\end{center}
\end{table}
\setlength{\tabcolsep}{1.4pt}
When only the generated samples are used to train the classifier, the accuracy is 82.6-96.6\%, the sensitivity reaches to 71.4-93.1\% and the specificity is always higher than 90\%, except AB. The mean values of accuracy, sensitivity and specificity are 84.42\%, 78.40\% and 97.63\% respectively, which indicate that the generated samples are valid. Further, we took the real CISLs data which was used to train the GAN (2 folds) and the same amount of non-sign samples to fine-tune the pre-trained classifier. When fine-tuning the neural network, the learning rate is decreased to 0.001.

It can be seen that the mean values of the accuracy and the sensitivity improve by about 6.6\% and 15.7\% respectively after fine-tuning with real samples, which are higher than those of the traditional method of data augmentation by 6.7\% and 12.8\%. CNN classifier with the large number of parameters, which is trained from the scratch on such small dataset is always easy to lead to overfitting. And we performed the experiment of training from the scratch, the accuracy is about 50\% and the sensitivity and the specificity are about 50\%, or one of the sensitivity and the specificity is 0 and the other is 1. The results predictably show that the model suffers from serious overfitting when CNN classifier is training from the scratch. Obviously, generating a large number of valid CISLs based on the CVAE-GAN, then using them to pre-train the CNN classifier can significantly reduce the risks of overfitting, and improve the performance of classification on small dataset.

\begin{figure}
\centering
\includegraphics[width=0.5\paperwidth]{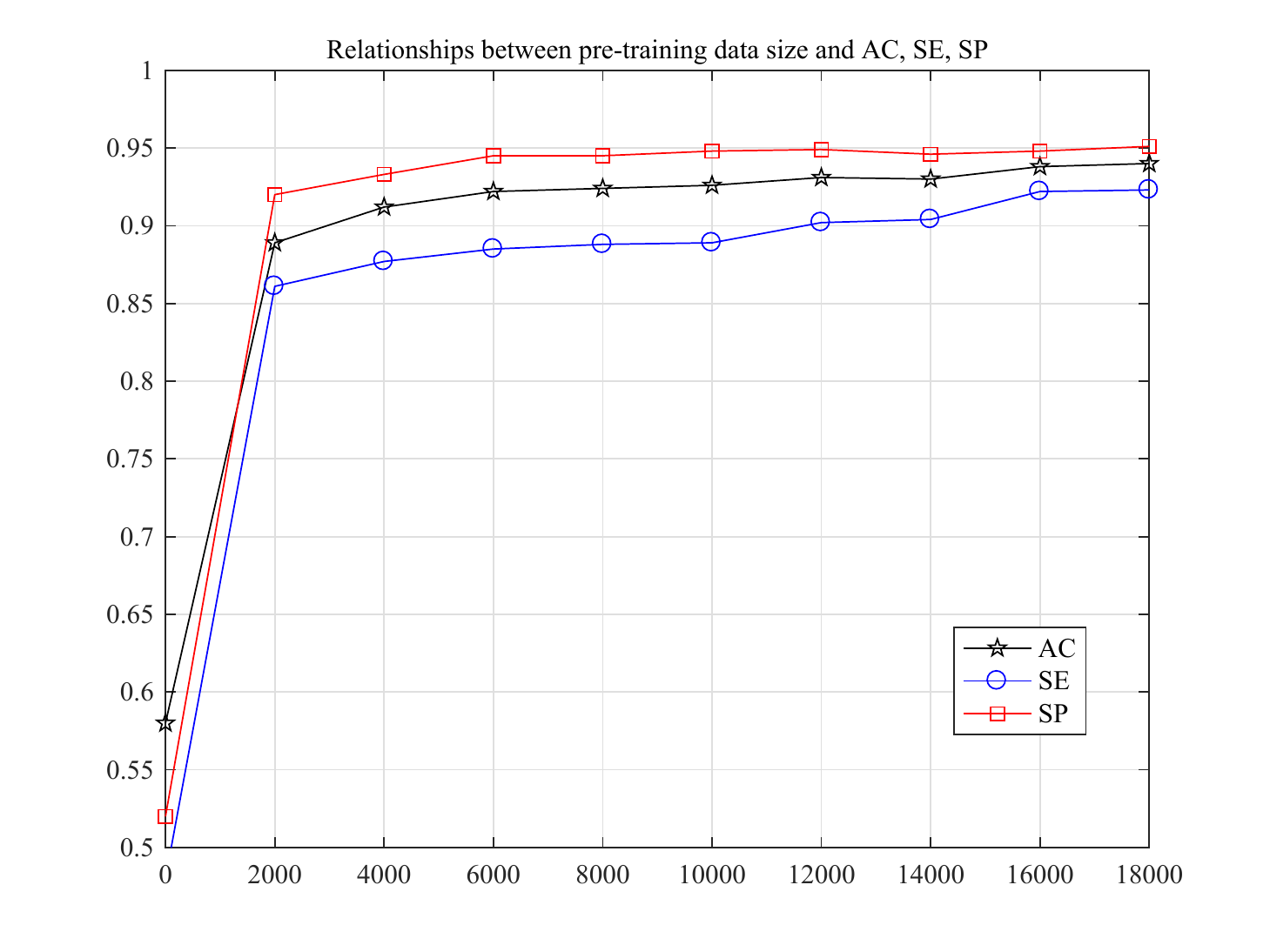}
\caption{The relationships between pre-training data size and the performance of binary classification for GGO after fine-tuning, accuracy (black), sensitivity (blue) and specificity (red).}
\label{fig:rel_augment}
\end{figure}

Fig.~\ref{fig:rel_augment} shows the relationship between the results of classification after fine-tuning and the number of generated GGO samples which were used to pre-train the classifier. As the number of samples increases, the strength of the classifier gradually increases. And when the number of samples rises to about 16000, the three indicators of classification performance reach the plateau. When the number of samples is more than 16000, the effect that increment of generated training data has on the results of classification becomes negligible.

\subsubsection{Multi-classification}
Next, we integrated 9 CISLs and normal tissues into 10 categories and trained CNN classifier for the task of 10-classification. For the sampling of normal tissues, we randomly cut normal regions from CT images which are not belonging to any types of CISLs. The number of samples of each category for pre-training the CNN model is 16k. The architecture and optimization algorithm of the CNN model are the same as those of the binary classification models except that the number of neurons in the last FC layer is changed to 10. And the accuracy, the sensitivity and the specificity are still used as indicators of classification performance. The classification results are shown in Table~\ref{table:table2}. 
As can be seen, when the classifier is learned from the scratch, the mean accuracy of 3-fold cross-validation is about 43.23\%, and the sensitivity of each type of CISLs is extremely low and the specificity is very high, but GGO is opposite. We analyze the reason and conclude that the number of GGO is more than the number of each type of other CISLs, the classifier trained from scratch is overfitting so that the model divides all CISLs into GGO for improving the global accuracy as much as possible. For the pre-trained model, the mean values of the accuracy, the sensitivity, and the specificity are about 79.65\%, 78.73\% and 97.44\%, respectively. After the neural net is fine-tuned, the mean values of the accuracy, the sensitivity and the specificity rise to about 91.83\%, 92.73\% and 99.0\%, and are 11.31\%, 22.54\% and 1.31\% higher than the traditional method. These results suggest that the method we proposed can effectively reduce the phenomenon of overfitting and is more effective than the traditional method of data augmentation for classification on small dataset. We confidently suggest that if the traditional scheme and proposed method are used together, we can get better results of classification.

\setlength{\tabcolsep}{1.5pt}
\begin{table}
\begin{center}
\caption{Sensitivity and specificity of 10-classification results.}
\label{table:table2}
\begin{tabular}{ccccccccccccc}
\hline\noalign{\smallskip}
\multirow{2}{*}{CISLs} & \multicolumn{3}{c}{Scratch} & \multicolumn{3}{c}{Pre-training} & \multicolumn{3}{c}{Fine-tuning} & \multicolumn{3}{c}{Traditional Method}\\
\cline{2-13}\noalign{\smallskip}
  & AC & SE & SP & AC & SE & SP & AC & SE & SP & AC & SE & SP\\
\noalign{\smallskip}
\hline
\noalign{\smallskip}
GGO  & \multirow{11}{*}{0.432} & 0.952 & 0.329
& \multirow{11}{*}{0.797} & 0.730 & 0.985 
& \multirow{11}{*}{0.927} & 0.919 & 0.974
& \multirow{11}{*}{0.805} & 0.958 & 0.971\\
lobulation &  & 0.333 & 0.998 &  & 0.896 & 0.998 &  & 0.972 & 0.998 &  & 0.807 & 0.980\\
calcification &  & 0.271 & 0.997 &  & 0.826 & 0.997 &  & 0.900 & 0.994 &  & 0.709 & 0.987\\
CV &  & 0.367 & 0.988 &  & 0.797 & 0.990 &  & 0.921 & 0.986  &  & 0.945 & 0.991\\
spiculation &  & 0.238 & 0.994 &  & 0.720 & 1.000 &  & 0.878 & 0.997  &  & 0.422 & 0.994\\
PI &  & 0.267 & 0.998 &  & 0.875 & 0.998 &  & 0.963 & 0.999  &  & 0.901 & 0.981\\
BMP &  & 0.197 & 0.988 &  & 0.420 & 0.997 &  & 0.852 & 0.986  &  & 0.674 & 0.936\\
AB &  & 0.278 & 0.998 &  & 0.909 & 0.998 &  & 1.000 & 0.998  &  & 0.274 & 0.990\\
OP &  & 0.200 & 1.000 &  & 0.714 & 1.000 &  & 0.952 & 1.000  &  & 0.500 & 0.995\\
normal tissue &  & 0.255 & 0.978 &  & 0.986 & 0.781 &  & 0.916 & 0.968  &  & 0.791 & 0.956\\
mean &  & 0.336 & 0.927 &  & 0.787 & 0.974 &  & 0.927 & 0.990  &  & 0.702 & 0.977\\
\hline
\end{tabular}
\end{center}
\end{table}
\setlength{\tabcolsep}{1.4pt}
\subsubsection{Comparing with Transfer Learning}
\begin{figure}
\centering
\subfigure[binary classification]{
	\includegraphics[width=0.4\paperwidth]{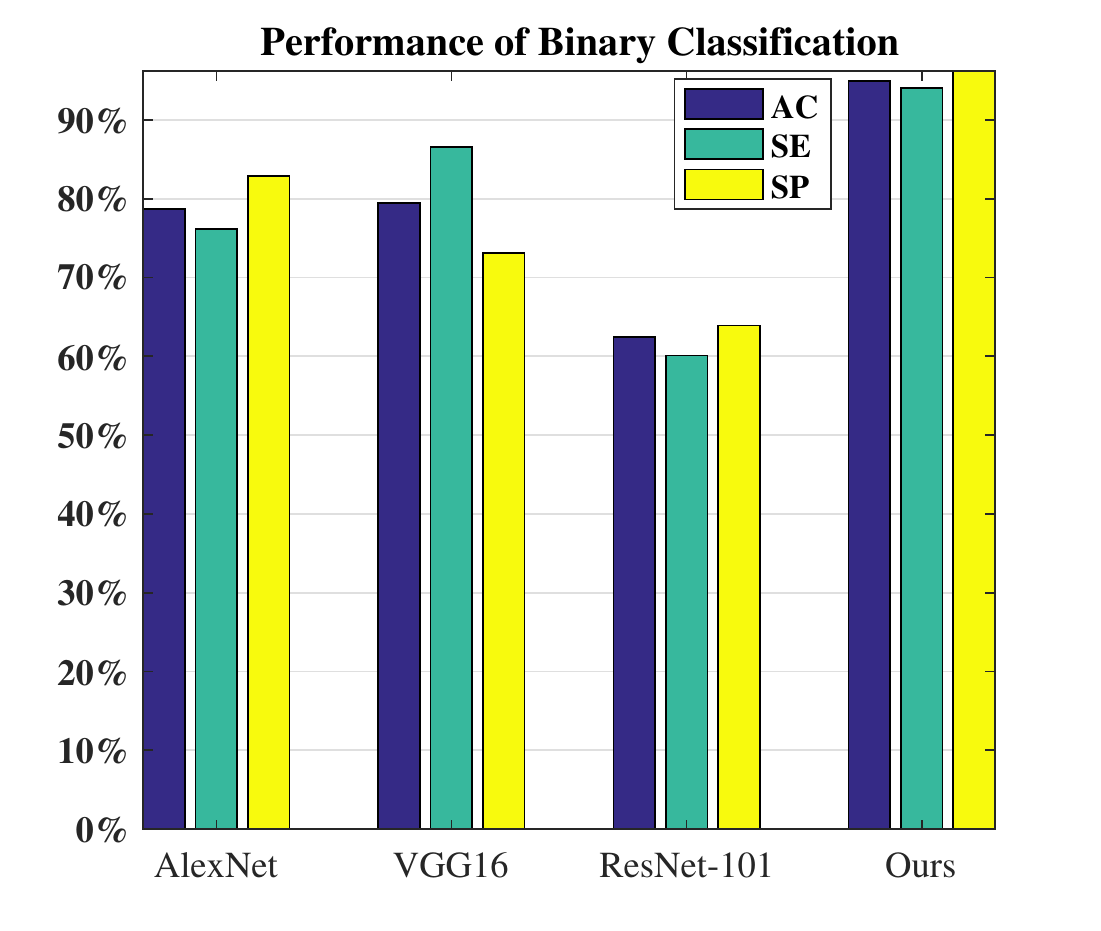}
	\label{fig:transfer:a}
}
\subfigure[multi-classification]{
	\includegraphics[width=0.4\paperwidth]{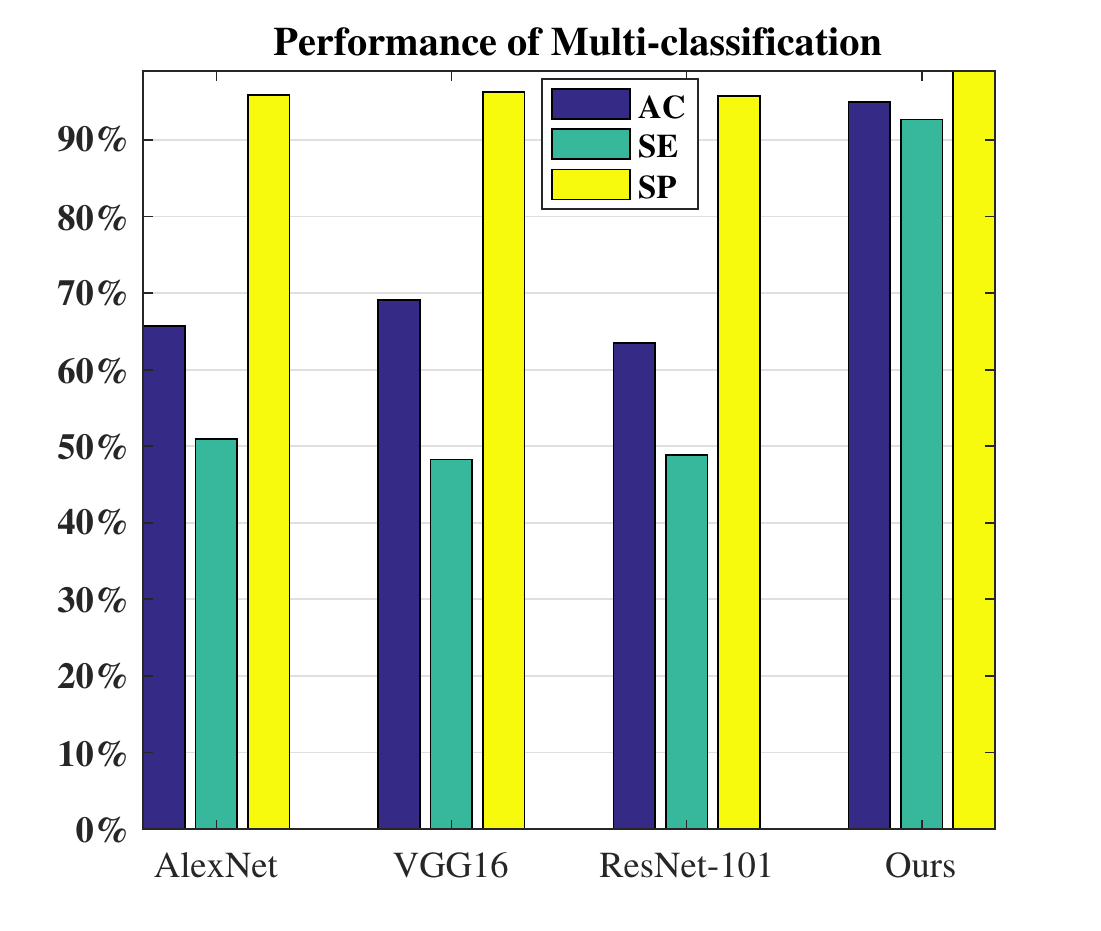}
	\label{fig:transfer:b}
}
\caption{The comparison between the results from the transfer learning with AlexNet, VGG16, ResNet-101 and those from our proposed method: (a) binary classification, (b) multi-classification.}
\label{fig:transfer}
\end{figure}

We compared the proposed method with several state-of-the-art CNN models which are AlexNet \cite{krizhevsky2012imagenet}, VGG16 \cite{simonyan2014very}, ResNet-101 \cite{he2016deep}. The three models are pre-trained on ImageNet dataset. As these models are trained with color images, we turn the single channel samples to three channels by duplicating the data of original channel. For AlexNet, we resize the samples to 227$\times$227 by the nearest interpolation function before feeding them into the net. For VGG16 and ResNet-101, we resize the samples to 224$\times$224. We always choose the last several layers to be fine-tuned and freeze the other layers. In experiments, we fine-tune the last three fully connected layers of AlexNet, the last two fully connected layers of VGG16 and the 4th block and logits layer of ResNet-101, respectively. We use the Adam optimizer and set the Hyperparameters of the epochs, batch size and learning rate are 20, 4 and 0.001 for these three models when fine-tuning. The experiment results are shown in Fig.~\ref{fig:transfer}.

Fig.~\ref{fig:transfer} shows that we achieve significant performance compared with fine-tuning three state-of-the-art models pre-trained on ImageNet dataset. VGG16 gains the best performance in three transfer learning models, and our method evidently outperforms the VGG16. We find that to some extent models which are trained with ImageNet dataset can provide rich features for medical imaging methods, but it is difficult to fine-tune these models on limited database due to the sophisticated architectures of neural nets, the risks of overfitting and the gap of cross modality. And we use relatively simple CNN models and no additional information from other datasets to achieve better results. These results demonstrate that our method can effectively reduce the chances of overfitting and improve the performance of CISLs classification.

\section{Conclusions}
In this paper, we have proposed a learning method of CNN for CISLs classification over small data, in which the CVAE-GAN is applied to generate samples, and the two-stages training scheme is employed to improve the performance of CNN classifier, i.e., these generated samples are used to pre-train the CNN classifier, then samples from target dataset are used for fine-tuning the classifier. We presented the quality of the generated samples and the classification performance. The results show that the generated samples have high quality, the radiologists averagely misidentified 56.7\% of the generated CISLs as real samples. The mean accuracy of binary classification and multi-classification achieve 95.0\% and 91.83\%, respectively. And our method achieves better performance compared with methods of traditional data augmentation and transfer learning from ImageNet to limited CISLs database. To sum up, the proposed method provides new scheme for deep classification methods on small datasets of medical images.

\end{document}